\documentclass[11pt, a4paper, logo, copyright]{eai}
\usepackage{shlab}

\usepackage[authoryear, sort&compress, round]{natbib}
\usepackage[]{mdframed}

\usepackage{dblfloatfix}

\usepackage{amsmath,amsfonts,bm}
\usepackage{multirow}
\usepackage{subcaption}
\usepackage{wrapfig}

\def\eqref#1{equation~\ref{#1}}

\def\1{\bm{1}}

\DeclareMathAlphabet{\mathsfit}{\encodingdefault}{\sfdefault}{m}{sl}
\SetMathAlphabet{\mathsfit}{bold}{\encodingdefault}{\sfdefault}{bx}{n}

\newcommand{\modelname}{InternVLA-A1.5\xspace}

\usepackage{listings}
\usepackage{hyperref}
\usepackage{url}
\usepackage{graphicx}
\usepackage{amsmath} %
\usepackage{mathrsfs} %
\usepackage{etoolbox}
\usepackage{cleveref}
\usepackage{tcolorbox}
\usepackage{colortbl}
\usepackage{booktabs}       %
\usepackage{amsfonts}       %
\usepackage{nicefrac}       %
\usepackage{microtype}      %
\usepackage{caption}
\usepackage{xspace} %
\usepackage{subcaption}
\usepackage{algorithm}
\usepackage{algorithmic}
\usepackage{multirow}
\usepackage{lipsum}

\usepackage{adjustbox}

\usepackage{booktabs} %
\usepackage{enumitem}%
\setlist[itemize]{noitemsep, topsep=0pt}
\usepackage{enumitem,kantlipsum}

\usepackage{multicol,multirow} 
\usepackage{graphicx}    
\usepackage{pifont}
\usepackage{minted}
\usepackage{ulem}
\usepackage{xcolor}
\usepackage{longtable}
\usepackage{makecell} 
\usepackage{listings}
\usepackage{xcolor}
\usepackage[T1]{fontenc}
\usepackage{placeins}

\lstdefinestyle{yamlstyle}{
    basicstyle=\ttfamily\footnotesize,
    numbers=left,
    numberstyle=\tiny,
    stepnumber=1,
    numbersep=6pt,
    frame=single,
    framerule=0.5pt,
    breaklines=true,
    breakatwhitespace=true,
    tabsize=2,
    captionpos=b,
    keywordstyle=\color{blue},
    commentstyle=\color{gray},
    stringstyle=\color{teal},
}

\newlength\savewidth

\definecolor{baselinecolor}{HTML}{d6eaf8}

\definecolor{mygray}{gray}{0.4}

\AtBeginEnvironment{tcolorbox}{\tiny}

\newcount\Comments  %
\usepackage{color}
\definecolor{darkred}{rgb}{0.9,0,0}
\definecolor{darkgreen}{rgb}{0,0.5,0}
\definecolor{darkblue}{rgb}{0,0,0.7}
\definecolor{purple}{rgb}{.6, 0,.6}
\definecolor{orange}{rgb}{1.0,0.64,0}

\definecolor{deemph}{gray}{0.6}

\definecolor{baselinecolor}{gray}{.9}
\definecolor{yellow}{RGB}{218,165,32}
\definecolor{lightcyan}{rgb}{0.88, 1.0, 1.0}
\definecolor{lightskyblue}{rgb}{0.53, 0.81, 0.98}
\definecolor{aliceblue}{rgb}{0.94, 0.97, 1.0}
\definecolor{LightSlateBlue}{RGB}{70,130,180}
\definecolor{DeepBlue}{RGB}{65,100,170}
\definecolor{DeepPurple}{RGB}{136,105,160}
\definecolor{LightGreen}{RGB}{59,125,35}
\definecolor{LightRed}{RGB}{234,66,53}
\definecolor{cvprblue}{rgb}{0.21,0.49,0.74}

\definecolor{mypink}{RGB}{254,102,140}
\definecolor{myclay}{RGB}{59,182,176}

\newcommand{\kibitz}[2]{\ifnum\Comments=1\textcolor{#1}{#2}\fi}

\title{InternVLA-A1.5: Unifying Understanding, Latent Foresight, and Action for Compositional Generalization}

\renewcommand{\today}{}

\author[]{Physical Intelligence Team, Shanghai AI Laboratory \\ {Full author list in \hyperref[sec:contributors]{Contributors} section}}

\renewcommand{\footnoterule}{%
  \kern -3pt
  \hrule width 1.0\columnwidth height 0.5pt
  \kern 2.6pt
}

\begin{document}

\begin{abstract}
\vspace{-10pt}
Unified models for robot manipulation aim to equip one policy with both the semantic priors of pretrained VLMs and the physical dynamics learned through future prediction.
In practice, existing designs tend to erode the semantics of the pretrained backbone, suffer interference among heterogeneous objectives, and learn future prediction from scratch in pixel space, leaving the dynamics priors of pretrained video generators unexploited. 
We present InternVLA-A1.5, which builds the policy on a native VLM backbone that keeps training on VQA and subtask prediction, and attaches a lightweight unified expert for continuous action generation. Future prediction is recast as a latent-querying problem, where a small set of learnable foresight tokens condenses the task-relevant future into a compact latent code under the supervision of a frozen pretrained video generation model, so the policy inherits world-model dynamics priors without ever learning pixel-level generation. The video branch is discarded at inference, keeping real-time control. 
Pretrained on 1.2M robot episodes and 3M multimodal samples, InternVLA-A1.5 achieves the best overall results on all six simulation benchmarks. In the real world, the preserved semantics deliver the strongest compositional generalization on held-out instruction bindings, and the two designs together sustain long-horizon execution.
\links{
  \link{homepage}{Homepage}{https://internrobotics.github.io/internvla-a15.github.io/}, 
  \link{code}{Code:InternVLA-A1.5}{https://github.com/InternRobotics/InternVLA-A-series}, 
  \link{huggingface}{Model:InternVLA-A1.5}{https://huggingface.co/collections/InternRobotics/internvla-a15},
}

\end{abstract}

\maketitle
\enlargethispage{3\baselineskip} 
\vspace{-18pt}
\begin{figure}[h]
    \centering
    \includegraphics[width=\textwidth]{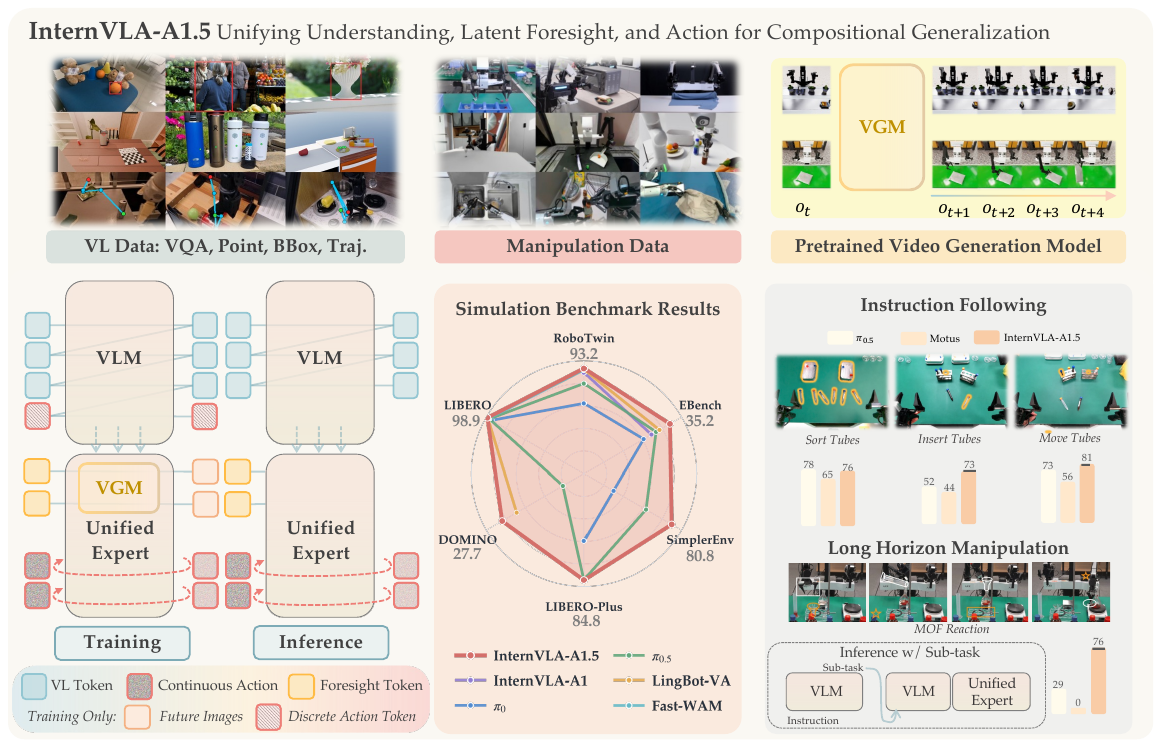} 
    \caption{\textbf{Overview of InternVLA-A1.5.} InternVLA-A1.5 unifies understanding, latent foresight, and action by attaching a lightweight expert to a pretrained VLM backbone. It is co-trained on vision-language and robot manipulation data, and introduces learnable foresight tokens to extract task-relevant future information as a compact latent representation. This representation is supervised by a frozen video generation model, which is used only during training and removed at inference. Extensive simulation and real-world experiments validate the effectiveness of the design.}
\label{fig:teaser}
\end{figure}

\section{Introduction}
\label{sec:introduction}

Developing general-purpose robots that can manipulate diverse objects under language instructions is a long-standing goal of embodied intelligence, with broad application value in homes, factories, and other unstructured environments. 
The strong generalization of vision-language models~\citep{achiam2023gpt,beyer2024paligemma,qwen3.5} and the powerful generative ability of video generation models~\citep{agarwal2025cosmos,wan2025wan} have motivated researchers to explore transferring such capabilities to embodied control. 
This has given rise to Vision-Language-Action (VLA) models~\citep{intelligence2024pi0,intelligence2025pi05,intelligence2025pi06,intelligence2026pi07,bjorck2025gr00t_n1,yu2026wall}, which inherit rich semantic priors from pretrained vision-language backbones and generalize well across objects and instructions.
In parallel, Video Action and World Action models~\citep{motus,li2026lingbot_va,kim2026cosmos_policy,dreamzero2025,team2026motubrain,zhou2026tau_0} learn to predict future visual states and thus acquire a strong sense of physical dynamics.
This complementarity has motivated a growing body of work to explore whether a single unified model~\citep{lu2025uniugp,hu2026bagelvla,liu2026mmada,cai2026internvla,sun2026vla,luo2026being} can combine the strengths of both. Along this line, our prior work InternVLA-A1~\citep{cai2026internvla} makes an early attempt by treating future visual states and actions jointly as training targets within a unified architecture, showing clear gains especially in dynamic manipulation.

Despite their effectiveness, current unified models that bring future prediction and action together still share several limitations rooted in how the two objectives are combined. 
First, the understanding component is often left out of large-scale VQA or language training once heavy generation and action objectives are imposed on top of it~\citep{lu2025uniugp,liu2026mmada,cai2026internvla,sun2026vla,luo2026being}. The semantic knowledge of the pretrained vision-language backbone then gradually drifts, and the instruction-following ability is weakened. 
Second, the sub-modules are optimized towards objectives of different forms and scales, for example regressing future visual latents, flow-matching action prediction, and language modeling, and these heterogeneous objectives tend to interfere with one another during joint training~\citep{luo2026being,sun2026vla,yu2026wall}. 
Third, the visual prediction module is usually trained from scratch to reconstruct future states, so it does not exploit the spatiotemporal and dynamics priors already contained in large pretrained video generation models~\citep{cai2026internvla,hu2026bagelvla}. 
Overall, these limitations reflect a common tension in unifying semantics and dynamics: how to inject knowledge of world dynamics into the policy without degrading its semantic ability or paying the full cost of pixel-level generation.

To address these limitations, we present InternVLA-A1.5. 
First, we build the policy on a native vision-language model~\citep{qwen3.5} and keep training it during policy learning with VQA, subtask prediction, and a discrete action token objective~\citep{intelligence2025pi05}. This preserves and even strengthens the semantic and instruction-following ability of the underlying VLM, while the discrete action objective provides an action-aware signal that speeds up convergence. 
Second, instead of training a generation module from scratch to regress future images in pixel space, we recast future prediction as a latent-querying problem: a small set of latent foresight queries attends to the shared multimodal context and reads out the task-relevant future as a compact latent code. This code is then used as the conditioning input of a frozen pretrained video generation model~\citep{wan2025wan}, and a video prediction loss is back-propagated through the frozen generator down to the foresight queries. 
In this way, the foresight queries are optimized to become a code that can steer a powerful generative world model toward the correct future, so that the policy inherits the model's spatiotemporal dynamics priors without ever learning pixel-level generation, and the auxiliary foresight objective is unified with action learning into a single and more consistent training framework. 
Notably, the video generation branch is used only during training and visualization; at deployment, action prediction does not invoke the generator, so the policy keeps a real-time inference speed.


We conduct extensive experiments in both simulation and the real world. In simulation, InternVLA-A1.5 is evaluated on six benchmarks, where LIBERO~\citep{liu2023libero}, RoboTwin 2.0~\citep{chen2025robotwin}, EBench~\citep{ebench2026}, and SimplerEnv~\citep{li2025evaluating} cover single-arm, bimanual, and mobile manipulation as well as real-to-sim evaluation, while LIBERO-Plus~\citep{fei2025libero} and DOMINO~\citep{fang2026towards} serve as zero-shot generalization tests. In the real world, we design three instruction-following tasks where only part of the instruction bindings are seen during training, together with a long-horizon chemistry procedure, and compare against $\pi_{0.5}$~\citep{intelligence2025pi05} and Motus~\citep{motus}. InternVLA-A1.5 achieves the best results on all six simulation benchmarks and stays ahead on both zero-shot tests. In the real world, it delivers the strongest compositional generalization on the held-out bindings and clearly leads in long-horizon execution.

\begin{figure}[t]
    \centering
    \includegraphics[width=\textwidth]{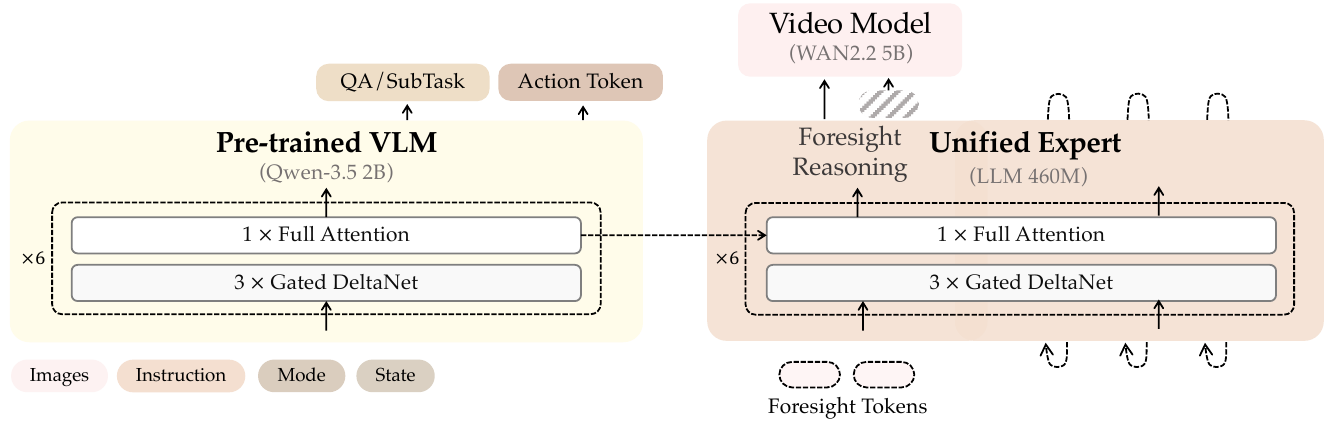} 
    \caption{
        \textbf{Framework of \modelname}. The architecture adopts a Mixture-of-Transformers design comprising a \textbf{pretrained VLM} for multimodal perception and a lightweight \textbf{unified expert} that shares full attention layers with the VLM while maintaining separate linear attention layers. The unified expert uses learnable foresight tokens for training-only visual foresight supervision and action query tokens decoded via flow matching for continuous action prediction.
    }
\label{figure:method_internvla}
\end{figure}

\section{InternVLA-A1.5 Model Design}
\label{sec:internvla-a1.5}

In this section, we present the design of \modelname, which adopts a Mixture-of-Transformers (MoT) architecture that unifies vision-language understanding, visual foresight,
and action generation within a single framework.
As illustrated in Figure~\ref{figure:method_internvla}, \modelname comprises two core components:
(1)~a pretrained VLM that serves as the backbone for multimodal perception and reasoning, and
(2)~a lightweight unified expert that shares the same architectural blueprint as the VLM backbone but operates with a smaller hidden dimension, enabling efficient action prediction.
We adopt Qwen-3.5 2B~\citep{qwen3.5} as the VLM backbone, which employs an efficient hybrid attention mechanism that interleaves 3~Gated DeltaNet~\citep{yang2025gated} linear
attention layers with 1~standard full attention layer.
The VLM and the unified expert only interact through the shared full attention layer, while maintaining separate Gated DeltaNet layers for modality-specific processing.

The VLM backbone is initialized from pretrained weights. At each timestep $t$, the multi-view observations $o_t$ from robot cameras and the language instruction $l$ are encoded into visual and text tokens following the standard VLM processing pipeline.
In addition, the robot proprioceptive state $q_t$ is discretized via uniform binning and appended as discrete tokens, along with a control mode token that specifies the action space.
This input formulation preserves the native representation of the pretrained VLM, allowing \modelname to fully leverage its multimodal understanding capability.
We employ a multi-stage training pipeline, with details described in Section~\ref{sec:training_recipe}. Depending on the task configuration, the VLM may predict textual outputs such as answers to visual questions or sub-task descriptions of the current state, as well as action chunks encoded as discrete tokens via the FAST tokenizer~\citep{pertsch2025fast}.

The unified expert adopts the same hybrid architecture as the VLM backbone (i.e., Qwen-3.5-Text) but with a reduced hidden dimension (460M parameters), and interacts with the VLM through the shared full attention layer to receive multimodal context.
The input to the unified expert consists of two parts. 
The first part is a set of learnable \textit{foresight tokens}
that serve as latent queries for future prediction. These tokens attend to the shared multimodal context via the
full attention layer and produce output embeddings that act as conditioning signals for a frozen pretrained video
generation model.
Specifically, we adopt WAN2.2-5B~\citep{wan2025wan} as the video generator and replace its original T5 text encoder
with the foresight query outputs. The model is supervised with a video prediction loss over future frames spanning the same temporal horizon as the action chunk.
Crucially, the video generation branch is used only during training; at inference time, the video model is discarded entirely, introducing no additional latency compared with world-model-based methods that require expensive generation at deployment.
Following the foresight tokens, a set of action query tokens is appended and decoded via a flow matching head to predict continuous action chunks. The action head can attend to all preceding non-action tokens.

\section{Training Recipe}
\label{sec:training_recipe}
The training of \modelname follows a multi-stage recipe.
The first pretraining stage co-trains the VLM backbone on a mixture of Vision-Question-Answer and large-scale robot data, in the spirit of recent works like $\pi_{0.5}$~\citep{intelligence2025pi05}.
The model is jointly supervised to predict the question answer, the next subtask, and a discrete action chunk.
This turns the VLM into a VLA executor while preserving its pretrained reasoning and instruction-following ability.
The second pretraining stage introduces the unified expert together with a foresight-reasoning mechanism.
A set of learnable foresight tokens condenses world knowledge from a frozen pretrained video generation model, injecting spatiotemporal priors into action prediction.
The post-training stage follows the same recipe as the second pretraining stage.
The video generation branch can be optionally kept active to fine-tune the foresight tokens on downstream demonstrations.

\subsection{Pretraining Stage 1: VLM Transferring}
\label{sec:stage1}

To make full use of the understanding capability carried from the pretrained VLM, we cast both robot data and VQA data into a single tokenized format and train the VLM under one unified framework.
The remainder of this subsection describes the encoding of each input modality and the prompt construction, the training objective, and the attention-mask layout.

\begin{figure}[t]
    \centering
    \includegraphics[width=\textwidth]{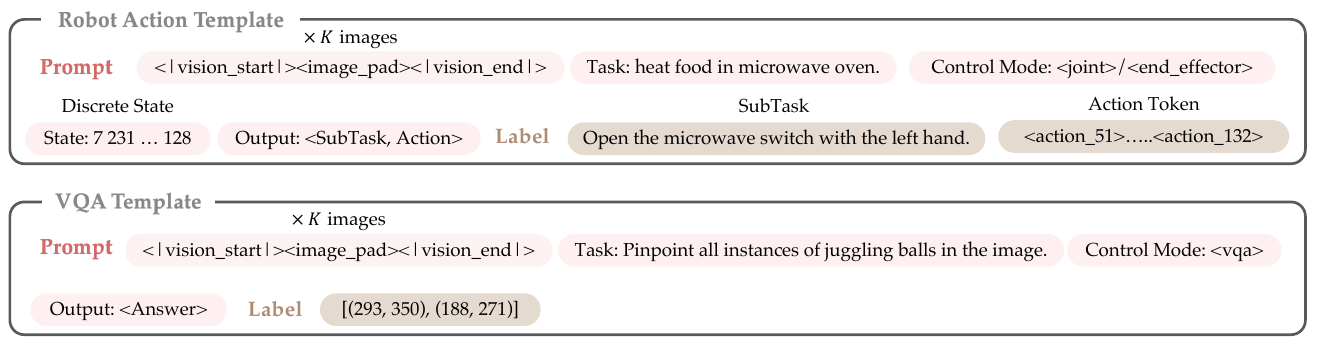}
    \caption{
        \textbf{The chat-template layout.}
        Robot action samples and VQA samples share the same prompt-and-label structure.
        The prompt concatenates $K$ image blocks, the task instruction, the control mode $m$, and the discretized state. The label part carries the supervised target: a subtask description and a FAST action token segment for robot data, or the answer span for VQA. Only label tokens contribute to the next-token cross-entropy.
    }
    \label{fig:train_chat_template}
\end{figure}

\paragraph{Input structure.}
At each timestep $t$, the input to the VLM consists of $K$-view observations $o_t = \{o_t^{(k)}\}_{k=1}^{K}$, a language instruction $l$, a control mode $m \in \{\textit{<joint>}, \textit{<end\_effector>}, \textit{<vqa>} \} $, and, for robot data, a proprioceptive state $q_t \in \mathbb{R}^{D}$ with $D \leq 32$.
Images are processed by the Qwen3.5 vision pipeline and inserted as $K$ blocks of \textit{<|vision\_start|>}\,\textit{<|image\_pad|>}\,\textit{<|vision\_end|>} tokens, with padded views masked out.
The proprioceptive state is uniformly discretized per dimension into $256$ bins over $[-1, 1]$.
Action targets are encoded as discrete tokens, where each fixed-horizon action chunk $a_{t:t+H}$ is mapped to a short sequence of tokens via the FAST tokenizer.
In practice, we adopt an action vocabulary of size $2048$ and append it to the original VLM vocabulary, so that action tokens share the same embedding table and language head as the rest of the model.
VQA and robot samples follow the unified chat-template layout shown in Figure~\ref{fig:train_chat_template}, which makes the prompt-versus-label split explicit.
For robot data, the prompt concatenates the image blocks, task instruction, control mode, and discretized state, and the label spans a subtask description followed by a FAST action segment, with a label-mode flag selecting which of the two (or both) is supervised per sample.
For VQA, the prompt drops the state segment and sets $m$ to \textit{<vqa>}, while the label collapses to a single answer span.

\paragraph{Training objective.}
The model is trained with a next-token prediction objective over the label portion of each sequence.
Following $\pi_{0.5}$~\citep{intelligence2025pi05}, we decompose the joint distribution of subtask output $\hat{\ell}$ and action chunk $\mathbf{a}_{t:t+H}$ as:
\begin{equation}
    \pi_\theta(\mathbf{a}_{t:t+H},\, \hat{\ell} \mid \mathbf{o}_t,\, \ell) = \pi_\theta(\mathbf{a}_{t:t+H} \mid \mathbf{o}_t,\, \hat{\ell})\; \pi_\theta(\hat{\ell} \mid \mathbf{o}_t,\, \ell),
\end{equation}
where $\mathbf{o}_t$ comprises multi-view images, instruction and the discretized proprioceptive state, and $\ell$ is the language instruction.
Since $\hat{\ell} = (\hat{\ell}_1, \dots, \hat{\ell}_M)$ is placed before the FAST action tokens $\mathbf{a} = (a_1, \dots, a_N)$ in the label sequence, the autoregressive factorization naturally conditions action prediction on the predicted subtask.
Denoting the full label sequence as $y = (y_1, \dots, y_{M+N}) = (\hat{\ell}_1, \dots, \hat{\ell}_M, a_1, \dots, a_N)$, the training objective is the standard cross-entropy loss over all label tokens:
\begin{equation}
    \mathcal{L}_{\text{stage1}} = -\mathbb{E}_{(\mathbf{o}_t,\,\ell,\,y)\,\sim\,\mathcal{D}} \left[ \sum_{i=1}^{M+N} \log p_\theta\!\left(y_i \mid \mathbf{o}_t, \ell, y_{<i}\right) \right],
\end{equation}
where $\mathcal{D}$ denotes the mixture of robot and VQA datasets.
For VQA samples, $\hat{\ell}$ reduces to the answer span and the action term vanishes.
Since action tokens are appended to the VLM vocabulary and share the same embedding table and output projection, all label tokens are supervised under a single cross-entropy loss without auxiliary heads or separate loss weighting.

\subsection{Pretraining Stage 2: Foresight and Action Generation}
\label{sec:stage2}

Although Stage~1 turns the VLM into a semantic-aware executor, its discrete autoregressive action prediction is inefficient for real-world closed-loop control, where low latency, precision, and smoothness are critical.
To overcome this limitation, Stage~2 pretraining introduces a unified expert that performs MoT joint attention with the VLM backbone.
The unified expert models the continuous action distribution and generates action chunks through flow-matching-based sampling.
Furthermore, to enhance the model's ability to reason about future scene states, we introduce a lightweight foresight-reasoning mechanism, where learnable foresight tokens query and absorb world-model knowledge from a frozen pretrained video generation model.
%
%
Overall, this stage retains the semantic and task-level reasoning ability acquired in Stage~1, while equipping the model with a future-aware, efficient, and control-oriented action generation interface.

\begin{figure}[ht]
    \centering
    \includegraphics[width=\textwidth]{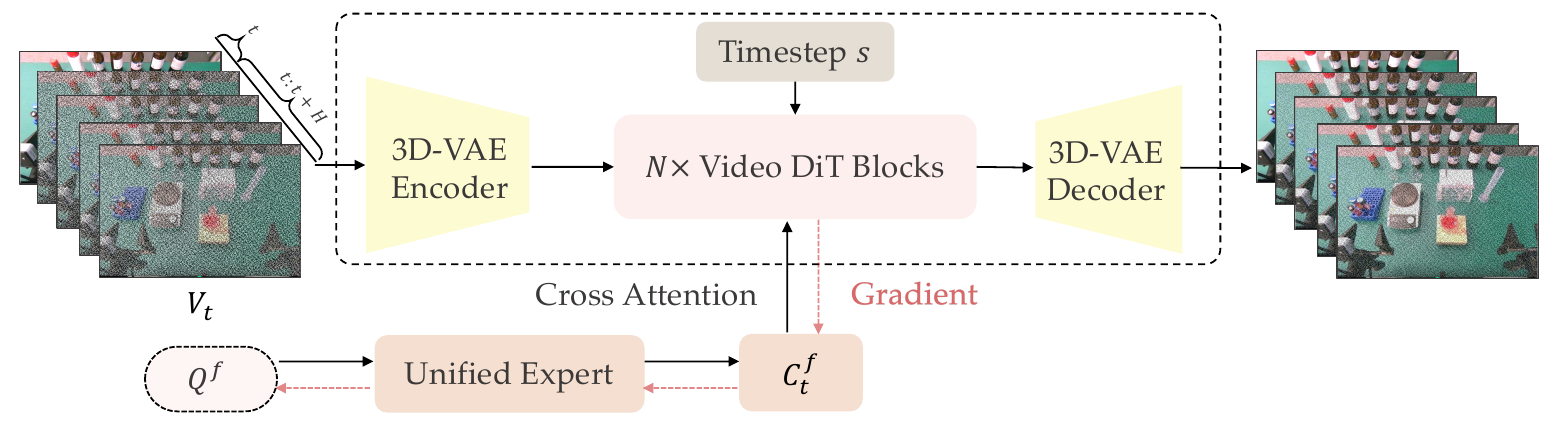}
    \caption{
        \textbf{The foresight reasoning mechanism.}
        Learnable foresight query tokens attend to the current visual-language context through the unified expert and produce conditioning embeddings for the frozen video generation model.
        The video supervision loss encourages these tokens to encode future-relevant information, which is further reused to guide continuous action generation.
    }
    \label{fig:foresight_reasoning}
\end{figure}

\paragraph{Foresight reasoning.}
Figure~\ref{fig:foresight_reasoning} illustrates the overall foresight reasoning pipeline and how learnable query tokens interact with the frozen video generation model. To leverage the dynamics priors encoded in pretrained video generation models, we insert $M$ learnable \textit{foresight tokens} into the unified expert sequence before flow-matching-based action prediction.
These tokens interact with the current visual-language context through the unified expert transformer, producing a compact representation that bridges future-state modeling and continuous action generation.
Denoting the learnable foresight tokens as $Q^f \in \mathbb{R}^{M \times d}$, the visual-language hidden states encoded from $(o_t,\ell,\hat{\ell})$ as $H_t$, the unified expert transformer as $\Phi_{\theta}$, and the foresight-token positions as $\mathcal{F}$, the contextualized foresight embeddings are obtained by
\begin{equation}
    Z_t^{f} = \Phi_{\theta}\!\left([H_t; Q^f]\right)_{\mathcal{F}}.
\end{equation}
The foresight embeddings are then projected into the conditioning space of the video generation model as $C_t^{f}=P_{\mathrm{WAN}}(Z_t^{f})$.

We instantiate the video generation model with a pretrained WAN2.2 model~\citep{wan2025wan} and freeze it during all training stages.
Instead of using the original T5 text encoder~\citep{2020t5}, we use $C_t^{f}$ as the condition and inject it through the native cross-attention layers of the WAN denoising transformer.
For each action chunk, we uniformly sample $N$ future frames as the prediction target, with $N=4$ in practice.
Denoting the concatenated current-and-future video clip as $V_t \in \mathbb{R}^{(1+N)\times H_I\times W_I\times 3}$, the WAN-VAE encoder compresses it into a clean video latent $x_1$.

Following the flow-matching objective used in WAN, we sample a noise latent $x_0 \sim \mathcal{N}(0,I)$ and an interpolation timestep $s \in [0,1]$.
Let $x_s=(1-s)x_0+s x_1$ be the interpolated latent and $v_s=x_1-x_0$ be the target velocity.
The video supervision loss is defined as
\begin{equation}
    \mathcal{L}_{\mathrm{video}} = \mathbb{E}_{x_0,x_1,C_t^{f},s}
    \left\|
    u(x_s,C_t^{f},s)-v_s
    \right\|^2,
\end{equation}
where $u$ denotes the frozen WAN denoising transformer.
Since the WAN parameters are frozen, gradients from $\mathcal{L}_{\mathrm{video}}$ are propagated only through the conditioning pathway, updating the foresight tokens $Q^{f}$ and the upstream unified expert layers that produce $C_t^{f}$.
As shown in Figure~\ref{fig:foresight_reasoning}, the foresight tokens function as learnable query slots that extract future-relevant signals from the current context, while the frozen WAN backbone provides the spatiotemporal prior used for supervision.
In this way, the foresight queries are trained to retrieve video-model-readable information from the current visual-language context, yielding future-aware embeddings that are further reused to condition continuous action generation.

\paragraph{Action prediction.}
In parallel with foresight prediction, the unified expert is trained to generate continuous action chunks with a flow-matching objective.
Unlike Stage~1, where actions are represented as discrete FAST tokens, Stage~2 directly models the continuous control trajectory, which is more suitable for low-latency closed-loop execution.
Given the visual-language context $H_t$ and the foresight query tokens $Q^f$, the unified expert predicts a velocity field that transports Gaussian noise to the expert action chunk.

Formally, let $\mathbf{a}_{t:t+H}$ denote the ground-truth continuous action chunk.
We sample Gaussian noise $\epsilon \sim \mathcal{N}(0,I)$ and an interpolation timestep $\tau \sim \mathrm{Beta}(1.5,1.0)$, and construct the interpolated action chunk
\begin{equation}
    \mathbf{a}_{t:t+H}^{\tau} = (1-\tau)\epsilon + \tau \mathbf{a}_{t:t+H}.
\end{equation}
The target velocity is therefore $\mathbf{a}_{t:t+H}-\epsilon$.
The action prediction loss is defined as
\begin{equation}
    \mathcal{L}_{\mathrm{action}}
    =
    \mathbb{E}_{\mathbf{a}_{t:t+H},\,\epsilon,\,\tau}
    \left\|
    v_{\theta}^{\mathrm{act}}
    \!\left(
        \mathbf{a}_{t:t+H}^{\tau}, H_t, Q^{f}
    \right)
    -
    \left(\mathbf{a}_{t:t+H}-\epsilon\right)
    \right\|^2,
\end{equation}
where $v_{\theta}^{\mathrm{act}}$ is the velocity field predicted by the unified expert.
The foresight tokens $Q^{f}$ serve as future-aware signals, enabling the unified expert to generate actions not only from the current observation and instruction, but also from the video-model-aligned representation of plausible future scene evolution.

During inference, we initialize the action chunk from Gaussian noise, $\mathbf{a}_{t:t+H}^{0}\sim\mathcal{N}(0,I)$, and solve the learned flow with Euler integration:
\begin{equation}
    \mathbf{a}_{t:t+H}^{\tau+\Delta \tau}
    =
    \mathbf{a}_{t:t+H}^{\tau}
    +
    \Delta \tau \cdot
    v_{\theta}^{\mathrm{act}}
    \!\left(
        \mathbf{a}_{t:t+H}^{\tau}, H_t, Q^{f}
    \right),
\end{equation}
where $\tau$ is advanced from $0$ to $1$ using $K$ integration steps with $\Delta \tau = 1/K$.
The final sample $\mathbf{a}_{t:t+H}^{1}$ is used as the predicted continuous action chunk.


\paragraph{Training objective.}
The Stage~2 objective combines the retained VLM next-token supervision with the video supervision and action flow-matching losses:
\begin{equation}
    \mathcal{L}_{\mathrm{stage2}}
    =
    \mathcal{L}_{\mathrm{stage1}}
    +
    \alpha \mathcal{L}_{\mathrm{video}}
    +
    \beta \mathcal{L}_{\mathrm{action}} .
\end{equation}
Here, $\mathcal{L}_{\mathrm{stage1}}$ denotes the same VLM cross-entropy objective used in Stage~1, while $\mathcal{L}_{\mathrm{video}}$ and $\mathcal{L}_{\mathrm{action}}$ supervise foresight conditioning and continuous action generation, respectively.
The weights $\alpha$ and $\beta$ balance the three loss terms; in practice, we set $\alpha=1$ and $\beta=10$.

\subsection{Attention Masking Pattern}
\label{sec:attention_mask}

\begin{figure}[ht]
    \centering
    \includegraphics[width=0.75\textwidth]{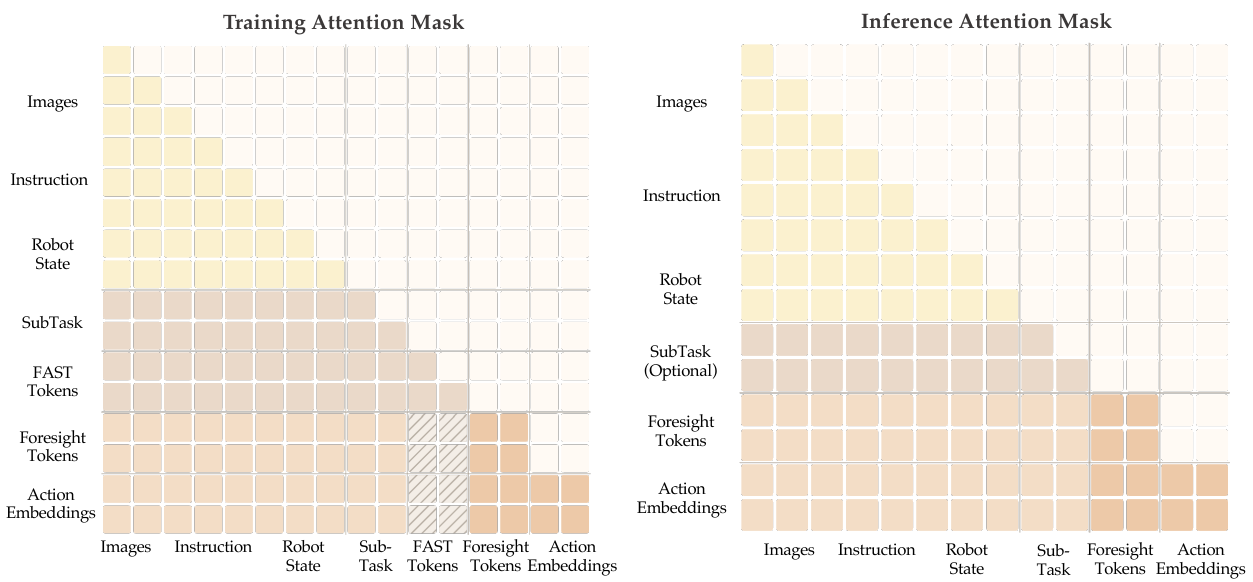}
    \caption{
        \textbf{Attention masking pattern of \modelname during training and inference.}
        The VLM tokens follow the causal attention pattern of Qwen3.5.
        The unified expert uses group-wise causal attention across token groups and bidirectional attention within each group.
        During training, the unified expert is prevented from attending to FAST action tokens to avoid information leakage and gradient interference.
    }
    \label{fig:attention_mask}
\end{figure}

As shown in Figure~\ref{fig:attention_mask}, \modelname uses a hybrid attention masking pattern to jointly support autoregressive language modeling and flow-matching-based action generation.
For the VLM part, including image tokens, instruction tokens, robot-state tokens, and the optional subtask tokens, we follow the original Qwen3.5 causal attention pattern.
This preserves the standard next-token prediction behavior used for subtask generation and, during training, for FAST action-token supervision.
For the unified expert, the learnable foresight query tokens and the noisy action embeddings are organized as separate token groups.
Across groups, we use a causal order: the foresight tokens attend to the VLM context, while the noisy action embeddings attend to both the VLM context and the foresight tokens.
Within each action-expert group, we use bidirectional attention, allowing the foresight tokens to exchange information with each other and the action embeddings to jointly model the whole action chunk.
This masking pattern matches the non-autoregressive nature of flow matching, where the entire noisy action chunk is denoised in parallel.

During training, FAST action tokens are present in the sequence for the discrete autoregressive objective, but we explicitly mask the attention from the unified expert to the FAST token span.
This prevents the continuous action generator from directly reading the ground-truth discrete action tokens and also avoids gradient interference between the FAST-token prediction branch and the flow-matching action branch.
For position ids, the unified expert tokens are assigned as if they were appended immediately after the optional subtask span, rather than after the FAST token span.
This keeps the positional layout of the unified expert consistent between training and inference.
During inference, FAST action tokens are not decoded. The unified expert reuses the KV cache of the VLM context during the denoising process, so repeated flow-sampling steps only need to update the denoising-dependent action computations.

\subsection{Training Protocol and Hyperparameters.}
During pretraining, we use the AdamW optimizer with a constant learning rate of $5\times10^{-5}$ for 300K steps in Stage 1 and 600K steps in Stage 2. Both stages use a batch size of 1024, a warmup of 2,000 steps, and a weight decay of 0.01. Gradient clipping is set to 1.0, and all models are trained in bfloat16 precision.
For post-training, we use a smaller batch size of 128 and apply the AdamW optimizer with a cosine decay schedule, where the learning rate decays from $5\times10^{-5}$ to $5\times10^{-6}$ over 60K steps. The warmup is set to 2,000 steps, while weight decay and gradient clipping remain unchanged.

Detailed hyperparameter configurations are summarized in Table~\ref{tab:training_hyperparameters}. The number of foresight tokens is set to $50$.

\begin{table}[h]
\centering
\caption{Training hyperparameters for \modelname.}
\small
\begin{tabular}{l c c c}
\toprule
\textbf{Configuration} & \textbf{Stage1 Pretrain} & \textbf{Stage2 Pretrain}& \textbf{Posttrain} \\
\midrule
Optimizer & AdamW &AdamW & AdamW \\
Batch size & 1024 & 1024 & 128 \\
Learning rate & $5 \times 10^{-5}$ & $5 \times 10^{-5}$ & $5 \times 10^{-5} \rightarrow 5 \times 10^{-6}$ \\
Warmup steps & 2,000 & 2,000 & 2,000 \\
Decay steps & -- & -- & 60,000 \\
Training steps & 300,000 & 600,000 & 60,000 \\
Foresight tokens & -- & 50 & 50 \\
Action chunk & 50 & 50 & 50 \\
Weight decay & 0.01 & 0.01 &0.01 \\
Gradient clipping & 1.0 &1.0 & 1.0 \\
Model precision & bfloat16 & bfloat16 & bfloat16 \\
\bottomrule
\end{tabular}
\label{tab:training_hyperparameters}
\end{table}

\section{Data Recipe}
\label{sec:data}

InternVLA-A1.5 is trained on two complementary data streams that mirror its
two learning objectives. A large-scale robot manipulation corpus supplies both
the action supervision and the future observations used for latent foresight,
while a multimodal corpus preserves the semantic and spatial-grounding ability
of the VLM backbone. We describe each stream below and then specify the
sampling strategy used to combine their heterogeneous sources.

\subsection{Robot Manipulation Data}
\label{subsec:robot_data}

\begin{figure*}[t]
\centering
\includegraphics[width=\textwidth]{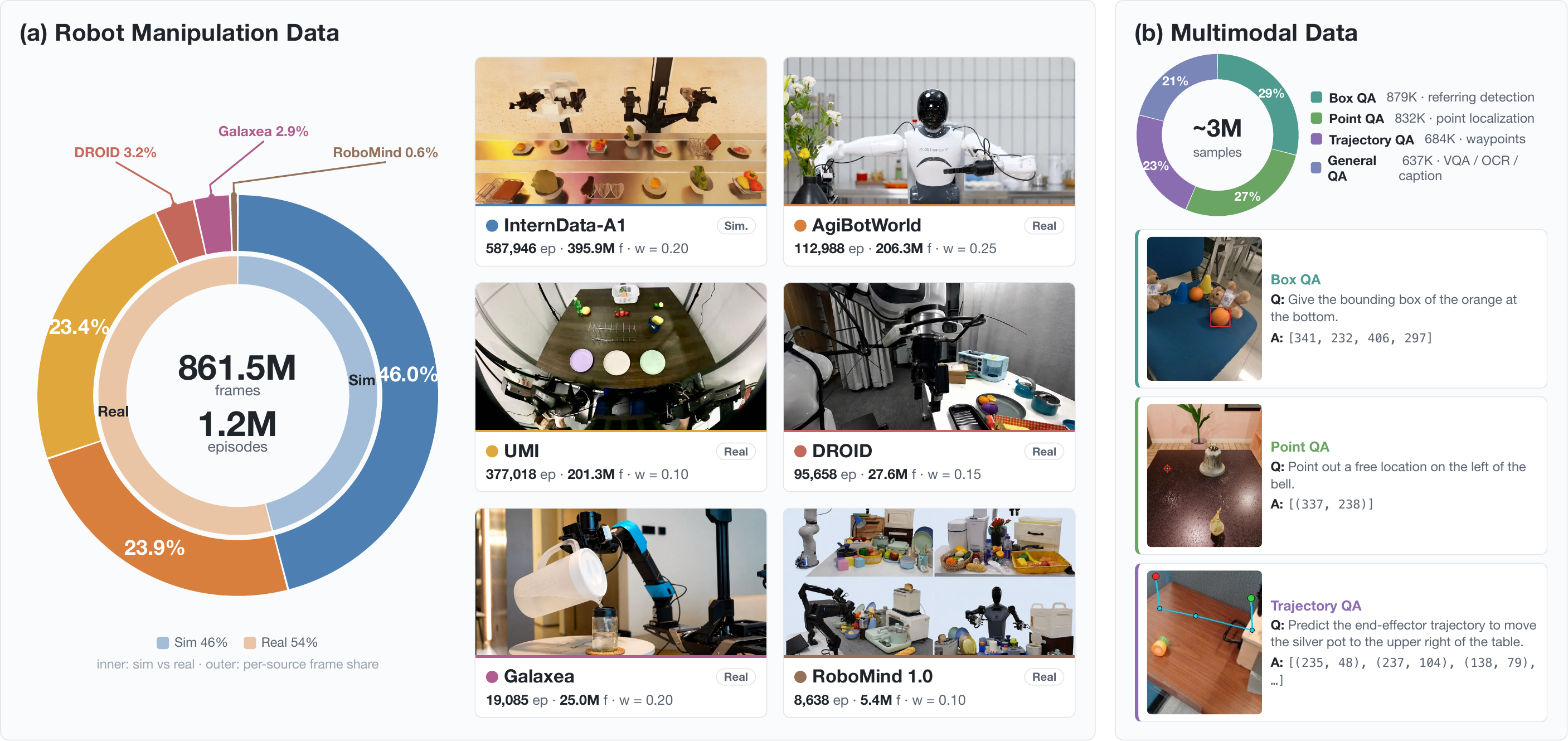}
\caption{\textbf{Overview of the training data for InternVLA-A1.5.}
\textbf{(a) Robot manipulation data.} Six sources, one synthetic
(InternData-A1) and five real-world, are unified into a single action space
and together provide 1.2M episodes and 861M frames. Each card reports the
episode count, frame count, and sampling weight of a source, and the outer
ring shows its share of frames. By frame count the corpus is split roughly
evenly between simulation and real-world data, while the sampling weights
shift the effective mixture toward the real-world sources.
\textbf{(b) Multimodal data.} The InternVLA-M1 corpus contributes
about 3M samples across four QA categories, one for general vision-language
ability and three for robotics-oriented spatial grounding, with representative
examples shown on the right.}
\label{fig:data_overview}
\end{figure*}

The robot corpus aggregates six sources, one synthetic and five real-world,
totaling 1.2M episodes and 861M frames (Figure~\ref{fig:data_overview}.a).
InternData-A1~\citep{tian2026interndata} provides the synthetic foundation and
the largest share of frames, with broad coverage of embodiments, skills, and
scenes. The real-world sources (AgiBotWorld~\citep{bu2025agibotworld},
UMI~\citep{jin2026umibench}, DROID~\citep{khazatsky2024droid},
Galaxea~\citep{jiang2025galaxea}, and RoboMind 1.0~\citep{wu2024robomind}) add
diversity in embodiments and viewpoints and help bridge the sim-to-real gap.
All sources are cast into the unified action space of InternVLA-A1~\citep{cai2026internvla}, with morphology-specific slots padded to a
shared layout so that every embodiment uses a single action head.

Each episode serves three roles during training. Its continuous action chunk
supervises the action prediction, and its future frames supervise the latent
foresight tokens through the frozen video generator. A FAST-tokenized form of
the same action additionally provides discrete targets for the VLM.

\subsection{Multimodal Co-training Data}
\label{subsec:vqa_data}

To keep the action and foresight objectives from eroding the VLM's pretrained
knowledge, we co-train on the multimodal corpus of
InternVLA-M1~\citep{chen2025internvla}, which pairs general vision-language
understanding with robotics-oriented spatial grounding. It contains about 3M
samples in four categories (Figure~\ref{fig:data_overview}.b). General QA (637K) maintains broad multimodal
ability, covering captioning, VQA, OCR, and knowledge grounding. The other
three are robotics-oriented grounding tasks, namely Box QA (879K) for referring
detection, Point QA (832K) for free-space and object-point localization, and
Trajectory QA (684K) for end-effector waypoint prediction. All grounding
targets use absolute image coordinates in a unified QA format, so they fit the
same next-token objective as the robot subtask and FAST targets. This stream
keeps the VLM's instruction-following and spatial-grounding ability intact and
available for the policy to exploit.

\subsection{Sampling Strategy}
\label{subsec:sampling}

The robot sources are imbalanced at two levels. Across sources, frame counts
span more than an order of magnitude, from 396M for InternData-A1 down to 5M
for RoboMind 1.0. Within each source, the distribution over its sub-datasets is
long-tailed. Proportional sampling on raw frame counts would therefore let a
few large sources dominate every batch and starve the smaller real-world
datasets that contribute most of the embodiment and scene diversity.

We therefore adopt a two-level grouped sampling scheme. Each source forms a
group, and within a group sub-datasets are drawn with probability proportional
to $(\#\mathrm{frames})^{\gamma}$. Setting $\gamma = 1$ recovers
frame-proportional sampling inside each group. The inter-group weights
(Figure~\ref{fig:data_overview}) are first obtained with Re-Mix~\citep{hejna2024re}
and then refined manually. The resulting weights up-weight the smaller
real-world sources (DROID, Galaxea, and RoboMind 1.0), whose raw frame share is
small but which add valuable embodiment and scene diversity, and down-weight
the dominant synthetic source. 

Finally, the robot corpus and the multimodal corpus are sampled at a fixed
0.15:0.85 ratio, so that the bulk of each batch reinforces the VLM's semantic
and grounding ability while the robot stream drives action and foresight
learning.

\section{Experiments}
\label{sec:experiments}

We conduct extensive experiments in both simulation and the real world to evaluate InternVLA-A1.5 from three perspectives, namely overall manipulation performance, training efficiency, and how effectively it exploits the dynamics knowledge of the pretrained world model. In simulation, we evaluate on six benchmarks. LIBERO~\citep{liu2023libero}, RoboTwin~\citep{chen2025robotwin}, EBench~\citep{ebench2026}, and SimplerEnv~\citep{li2025evaluating} together cover single-arm, bimanual, and mobile manipulation as well as real-to-sim evaluation. LIBERO-Plus~\citep{fei2025libero} and DOMINO~\citep{fang2026towards} serve as zero-shot generalization tests, where the model trained on LIBERO is evaluated directly under the camera, language, layout, and other perturbations of LIBERO-Plus, and the model trained on RoboTwin is evaluated directly on the dynamic manipulation tasks of DOMINO. In the real world, we design four tasks. The first three probe instruction following, where only part of the instruction bindings are seen during training, and the fourth is a multi-step chemistry procedure that targets long-horizon execution. We further examine the training efficiency and the subtask-following ability of InternVLA-A1.5, and analyze both qualitatively and quantitatively how the latent foresight helps the policy exploit the pretrained world model.

\subsection{Evaluation on Real-world Experiments}

\begin{figure}[h]
  \centering
  \includegraphics[width=\linewidth]{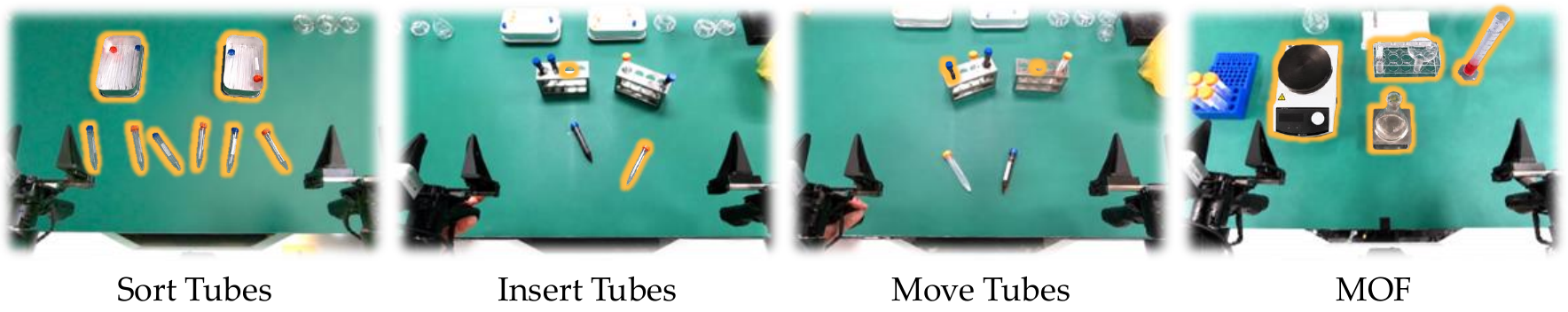}
  \caption{\textbf{Overview of the four real-world tasks.} The first three evaluate
  instruction following, where the robot brings a specified tube to a specified target,
  namely a box in Sort Tubes, a hole of a tube rack in Insert Tubes, and a position on the
  opposite side in Move Tubes. MOF is a long-horizon chemistry procedure. In each panel,
  colored outlines mark the task-relevant tubes and targets.}
  \label{fig:realworld-tasks}
\end{figure}

We evaluate InternVLA-A1.5 on four manipulation tasks (Figure~\ref{fig:realworld-tasks}). The first three target instruction following. In each, the robot follows a language instruction that names a specified tube and a specified target and must carry that tube to the target, and the three tasks differ only in the type of target, a box in Sort Tubes, a hole of a tube rack in Insert Tubes, and a position on the opposite side of the workspace in Move Tubes. 
All three tasks follow the same held-out design, where only part of the (tube, target) bindings are demonstrated during training, and the remaining bindings are evaluated only at test time, so that success requires grounding the tube and the target from language rather than replaying a memorized motion. The fourth task, MOF, is a long-horizon metal-organic framework preparation procedure that targets long-horizon execution.
Detailed setups for all four tasks are provided in Appendix~\ref{app:realworld-tasks}. All methods are fine-tuned on the same demonstrations and evaluated under an identical protocol, and for each evaluated condition we run multiple trials with randomized object placement and report the success rate (\%), comparing against Motus~\citep{motus} and $\pi_{0.5}$~\citep{intelligence2025pi05}. All real-world experiments run on a single NVIDIA RTX 5090 GPU. With static-graph execution, SDPA, and the flash linear attention library~\citep{yang2024fla}, one inference step of InternVLA-A1.5 takes about 0.1s. Since the video generation branch is discarded at deployment, the policy avoids the second-level per-step generation cost of world-action models that imagine future frames at test time, and supports real-time closed-loop control.

\begin{figure*}[t]
  \centering
  \includegraphics[width=0.65\linewidth]{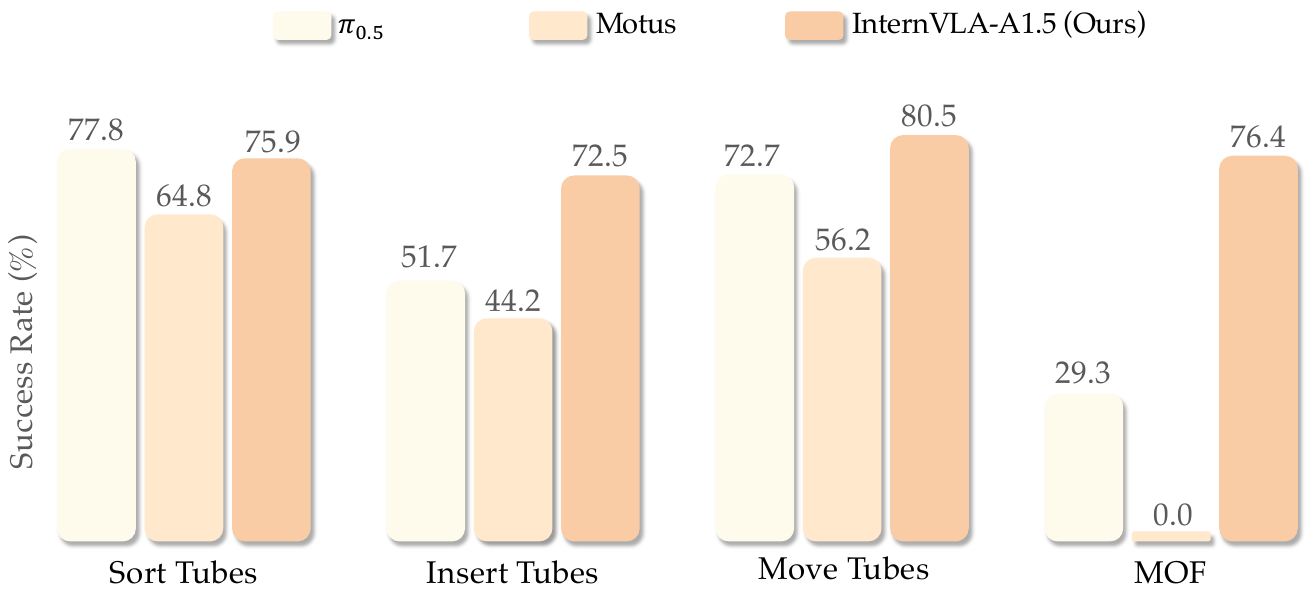}
  \caption{\textbf{Real-world results across instruction-following and long-horizon tasks.} We report overall success rates on Sort Tubes, Insert Tubes, Move Tubes, and MOF, comparing $\pi_{0.5}$, Motus, and \modelname.}
  \label{fig:real_exp_all}
\end{figure*}

\begin{figure*}[t]
  \centering
  \includegraphics[width=0.8\linewidth]{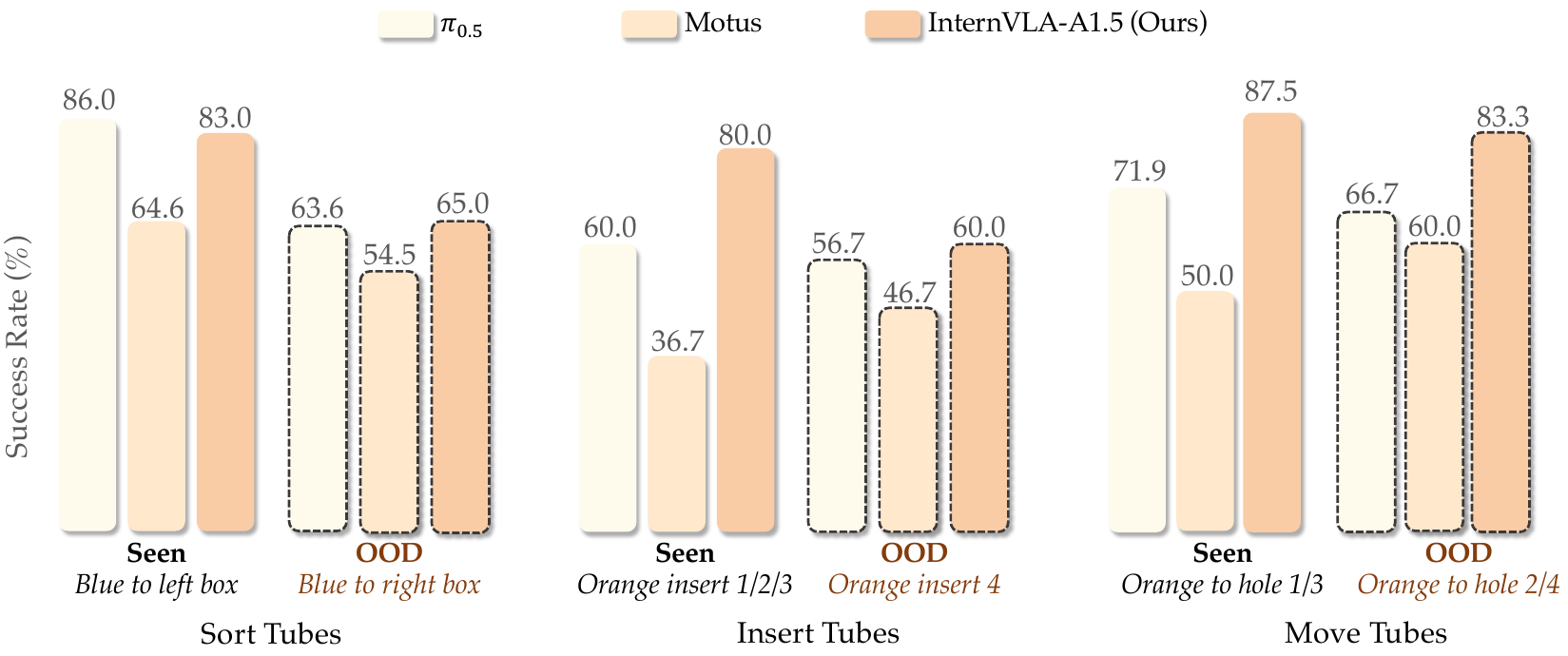}
  \caption{\textbf{Generalization to seen and out-of-distribution instruction bindings.} We decompose performance into seen vs. OOD combinations across the three instruction-following tasks.}
  \label{fig:real_exp_gen}
\end{figure*}

Figure~\ref{fig:real_exp_all} summarizes the overall real-world performance across all tasks. InternVLA-A1.5 improves over Motus on all four tasks and achieves the best success rate on Insert Tubes, Move Tubes, and MOF, while on Sort Tubes it is slightly behind $\pi_{0.5}$ (75.9 vs. 77.8). The three instruction-following tasks differ in the precision the target requires. Sort Tubes only asks the robot to drop the tube into an open box, which is close to a plain pick-and-place skill, and we observe that $\pi_{0.5}$ is already very strong on such skills. Insert Tubes and Move Tubes instead require inserting the tube into a specific hole of a rack, where InternVLA-A1.5 leads $\pi_{0.5}$ by 20.8 and 7.8 points respectively. On the long-horizon MOF task, InternVLA-A1.5 reaches 76.4\% while $\pi_{0.5}$ succeeds in only 29.3\% of the trials and Motus fails to complete the procedure in any trial, indicating a clear advantage in extended sequential execution. We attribute this gap to the explicit subtask prediction, which keeps the policy aware of task progress over long horizons, and to the learned dynamics priors, which let the policy reason about environment-changing actions such as liquid pouring without an explicit memory mechanism. $\pi_{0.5}$ also predicts subtasks but struggles to model such state transitions, and Motus lacks the subtask-level structure needed to track progress.

Figure~\ref{fig:real_exp_gen} decomposes the three instruction-following tasks into seen and held-out instruction bindings. On the held-out bindings, InternVLA-A1.5 obtains the best success rate on all three tasks, showing that its advantage does not come from replaying demonstrated bindings. This also puts the Sort Tubes result in context, as the small overall gap to $\pi_{0.5}$ comes mainly from the seen bindings, and on the held-out bindings that the task is designed to probe, InternVLA-A1.5 stays ahead. We also note that individual targets differ in physical difficulty, which is why some methods score higher on certain held-out bindings than on their seen ones, so the drop from seen to held-out partly reflects target difficulty rather than instruction grounding alone. Overall, InternVLA-A1.5 improves absolute success rates and remains robust under compositional shifts.

\begin{table*}[t]
\centering
\caption{Results on SimplerEnv (WidowX). All numbers are success rates (\%). Baseline numbers are compiled from prior studies that evaluate these models on this benchmark. The best and second-best results are highlighted in \textbf{bold} and \underline{underline}, respectively.}
\label{tab:simplerenv_widowx}
\resizebox{\linewidth}{!}{%
\begin{tabular}{lccccc}
\toprule
Method & \makecell{Put Spoon\\on Towel} & \makecell{Put Carrot\\on Plate} & \makecell{Stack Green Block\\on Yellow Block} & \makecell{Put Eggplant\\in Yellow Basket} & \textbf{Average} \\
\midrule
$\pi_0$~\citep{intelligence2024pi0} & 29.1 & 0.0 & 16.6 & 62.5 & 27.1 \\
$\pi_{0.5}$~\citep{intelligence2025pi05} & 49.3 & 64.7 & 44.7 & 69.7 & 57.1 \\
GR00T-N1.5~\citep{bjorck2025gr00t_n1} & 75.3 & 54.3 & 57.0 & 61.3 & 61.9 \\
InternVLA-M1~\citep{chen2025internvla} & 87.5 & \underline{67.9} & 31.3 & \textbf{100.0} & 71.7 \\
EO-1~\citep{qu2025eo} & 63.6 & 54.5 & \textbf{81.8} & \underline{90.9} & 72.7 \\
Xiaomi-Robotics-0~\citep{cai2026xiaomi} & \textbf{95.8} & 62.5 & \underline{75.0} & 83.3 & \underline{79.2} \\
\midrule
\textbf{InternVLA-A1.5 (Ours)} & \underline{92.4} & \textbf{85.4} & 69.5 & 75.7 & \textbf{80.8} \\
\bottomrule
\end{tabular}%
}
\end{table*}

\setcounter{topnumber}{3}
\setcounter{dbltopnumber}{3}
\setcounter{totalnumber}{5}
\renewcommand{\topfraction}{0.95}
\renewcommand{\dbltopfraction}{0.95}
\renewcommand{\textfraction}{0.05}
\setlength{\floatsep}{8pt plus 1pt minus 2pt}
\setlength{\dblfloatsep}{8pt plus 1pt minus 2pt}

\begin{table*}[t]
\centering
\begin{minipage}{0.49\linewidth}
\centering
\caption{Results on RoboTwin. We report success rates under clean and randomized evaluation settings, together with their average.}
\label{tab:robotwin}
\resizebox{\linewidth}{!}{%
\begin{tabular}{lccc}
\toprule
Method & Clean & Rand. & Avg. \\
\midrule
$\pi_0$~\citep{intelligence2024pi0} & 65.9 & 58.4 & 62.2 \\
$\pi_{0.5}$~\citep{intelligence2025pi05} & 82.7 & 76.8 & 79.8 \\
LingBot-VLA~\citep{wu2026pragmatic} & 86.5 & 85.3 & 85.9 \\
StarVLA~\citep{community2026starvla} & 88.7 & 87.8 & 88.3 \\
InternVLA-A1~\citep{cai2026internvla} & 89.4 & 89.6 & 89.5 \\
Motus~\citep{motus} & 88.7 & 87.0 & 87.8 \\
LingBot-VA~\citep{li2026lingbot_va} & \underline{92.9} & 91.5 & \underline{92.2} \\
Fast-WAM~\citep{yuan2026fast} & 91.9 & \underline{91.8} & 91.8 \\
Being-H0.7~\citep{luo2026being} & 90.2 & 89.6 & 89.9 \\
\midrule
\textbf{InternVLA-A1.5 (Ours)} & \textbf{93.3} & \textbf{93.0} & \textbf{93.2} \\
\bottomrule
\end{tabular}%
}
\end{minipage}
\hfill
\begin{minipage}{0.49\linewidth}
\centering
\caption{Comparison on DOMINO. SR is the primary metric, and MS denotes manipulation score. We mainly report the zero-shot settings; results of our fine-tuned model are also reported.}
\label{tab:domino}
\resizebox{\linewidth}{!}{%
\begin{tabular}{lcc}
\toprule
Method & SR (\%) $\uparrow$ & MS $\uparrow$ \\
\midrule
\multicolumn{3}{l}{\textit{Zero-shot to dynamic manipulation}} \\
\midrule
OpenVLA-OFT~\citep{kim2025fine} & 6.7 & 20.0 \\
$\pi_{0.5}$~\citep{intelligence2025pi05} & 7.5 & 20.4 \\
LingBot-VLA w/ depth~\citep{wu2026pragmatic} & 11.8 & 26.7 \\
LingBot-VA~\citep{li2026lingbot_va} & 24.1 & 36.1 \\
Qwen-VLA-Base~\citep{wang2026qwen} & 21.1 & 37.4 \\
Qwen-VLA-Instruct~\citep{wang2026qwen} & \underline{26.6} & \underline{39.5} \\
\midrule
\textbf{InternVLA-A1.5 (Zero-shot)} & \textbf{27.7} & \textbf{39.8} \\
\textbf{InternVLA-A1.5 (Fine-tuned)} & \textbf{29.3} & \textbf{42.5} \\
\bottomrule
\end{tabular}%
}
\end{minipage}
\end{table*}

\begin{table*}[t]
\centering
\caption{Results on LIBERO. The best and second-best results are highlighted in \textbf{bold} and \underline{underline}, respectively.}
\label{tab:libero}
\resizebox{0.7\linewidth}{!}{%
\begin{tabular}{lccccc}
\toprule
Method & Spatial & Object & Goal & Long & \textbf{Average} \\
\midrule
$\pi_0$~\citep{intelligence2024pi0} & 98.0 & 96.8 & 94.4 & 88.4 & 94.4 \\
$\pi_{0.5}$~\citep{intelligence2025pi05} & \textbf{98.8} & 98.2 & 98.0 & 92.4 & 96.9 \\
GR00T-N1.7~\citep{bjorck2025gr00t_n1} & 97.7 & 98.5 & 97.5 & 94.4 & 97.0 \\
OpenVLA-OFT~\citep{kim2025fine} & 97.6 & 98.4 & 97.9 & 94.5 & 97.1 \\
InternVLA-M1~\citep{chen2025internvla} & 98.0 & 99.0 & 93.8 & 92.6 & 95.9 \\
Xiaomi-Robotics-0~\citep{cai2026xiaomi} & \textbf{98.8} & \textbf{100.0} & \textbf{98.8} & 97.2 & \underline{98.7} \\
Motus~\citep{motus} & 96.8 & \underline{99.8} & 96.6 & 97.6 & 97.7 \\
LingBot-VA~\citep{li2026lingbot_va} & 98.5 & 99.6 & 97.2 & \textbf{98.5} & 98.5 \\
Fast-WAM~\citep{yuan2026fast} & 98.2 & \textbf{100.0} & 97.0 & 95.2 & 97.6 \\
\midrule
\textbf{InternVLA-A1.5 (Ours)} & \underline{98.6} & \underline{99.8} & \underline{98.6} & \underline{98.4} & \textbf{98.9} \\
\bottomrule
\end{tabular}%
}
\end{table*}

\begin{table*}[t]
\centering
\caption{Results on LIBERO-Plus. The best and second-best results are highlighted in \textbf{bold} and \underline{underline}, respectively.}
\label{tab:libero_ood}
\resizebox{\linewidth}{!}{%
\begin{tabular}{lcccccccc}
\toprule
Method & Camera & Robot & Language & Light & Background & Noise & Layout & \textbf{Total} \\
\midrule
$\pi_0$~\citep{intelligence2024pi0} & 13.8 & 6.0 & 58.8 & 85.0 & 81.4 & 79.0 & 68.9 & 53.6 \\
$\pi_{0.5}$~\citep{intelligence2025pi05} & \underline{78.4} & \textbf{73.6} & 80.8 & 96.2 & 94.1 & 89.0 & 84.5 & \underline{84.4} \\
StarVLA~\citep{community2026starvla} & 52.5 & 49.8 & \textbf{88.5} & 95.7 & \underline{95.7} & 73.0 & 76.9 & 74.1 \\
Abot-M0~\citep{yang2026abot} & 60.4 & \underline{67.9} & 86.4 & 96.2 & 91.6 & 86.4 & \underline{82.6} & 80.5 \\
Cosmos-Policy~\citep{kim2026cosmos_policy} & 75.8 & 63.3 & 81.7 & \textbf{96.5} & 88.9 & \underline{92.7} & 82.2 & 82.2 \\
\midrule
\textbf{InternVLA-A1.5 (Ours)} & \textbf{83.1} & 55.1 & \underline{86.9} & \underline{96.4} & \textbf{98.2} & \textbf{95.6} & \textbf{85.2} & \textbf{84.8} \\
\bottomrule
\end{tabular}%
}
\end{table*}

\begin{table*}[t]
\centering
\caption{
Results on EBench.
SR (\%) is the primary metric, Score denotes continuous task progress (\%). The best and second-best results are highlighted in \textbf{bold} and \underline{underline}, respectively.
}
\label{tab:ebench}
\resizebox{0.7\linewidth}{!}{%
\begin{tabular}{lcccccc}
\toprule
& \multicolumn{2}{c}{Val-Train} 
& \multicolumn{2}{c}{Val-Unseen} 
& \multicolumn{2}{c}{Test} \\
\cmidrule(lr){2-3}
\cmidrule(lr){4-5}
\cmidrule(lr){6-7}
Method 
& SR $\uparrow$ & Score $\uparrow$
& SR $\uparrow$ & Score $\uparrow$
& SR $\uparrow$ & Score $\uparrow$ \\
\midrule
$\pi_0$~\citep{intelligence2024pi0} 
& 30.5 & 42.9 
& 25.4 & 39.3 
& 24.4 & 38.4  \\

XVLA~\citep{zheng2025x} 
& 28.3 & 42.1 
& 22.7 & 35.9 
& 24.7 & 37.5  \\

InternVLA-A1~\citep{cai2026internvla} 
& 33.1 & 44.2 
& 20.8 & 33.8 
& 27.6 & 40.2  \\

$\pi_{0.5}$~\citep{intelligence2025pi05} 
& 32.1 & 48.1
& 26.5 & 42.9
& 29.5 & 45.6 \\

LingBot-VA~\citep{li2026lingbot_va}
& \underline{38.3} & \textbf{55.9} 
& \underline{26.6} & \underline{43.3}
& \underline{30.9} & \underline{47.6}  \\ \midrule

\textbf{\modelname (Ours)}
& \textbf{43.1} & \underline{54.1}
& \textbf{32.8} & \textbf{46.4} 
& \textbf{35.2} & \textbf{49.5}  \\
\bottomrule
\end{tabular}%
}
\end{table*}

\subsection{Evaluation on Simulation Benchmarks}

\paragraph{Benchmarks and setup.} We evaluate InternVLA-A1.5 on six simulation benchmarks that span different simulators, embodiments, and skill types. LIBERO~\citep{liu2023libero} uses a 7-DoF single arm and contains four suites (Spatial, Object, Goal, and Long), and we report the per-suite success rate and their average. LIBERO-Plus~\citep{fei2025libero} extends LIBERO with perturbations on camera, language, layout, and other factors, and we evaluate the LIBERO checkpoint on it in zero-shot without any further training. RoboTwin~\citep{chen2025robotwin} comprises 50 bimanual tasks under a clean and a domain-randomized setting. DOMINO~\citep{fang2026towards} builds dynamic manipulation tasks with moving objects on top of RoboTwin, and we evaluate the RoboTwin checkpoint on it zero-shot and additionally report a fine-tuned variant. EBench~\citep{ebench2026} is an indoor mobile manipulation benchmark built on Isaac Sim that covers long-horizon, pick-and-place, and dexterous skills. SimplerEnv~\citep{li2025evaluating} provides a real-to-sim evaluation on the WidowX (Bridge) setup, and we report the visual matching success rate. All fine-tuning starts from the same pretrained checkpoint and follows the standard data format and evaluation protocol of each benchmark, with benchmark-specific details given in Appendix~\ref{app:simbench}. Unless otherwise noted, we compare against the strong baseline $\pi_{0.5}$~\citep{intelligence2025pi05}, which is available on every benchmark, together with published baselines reported on each, including our prior model InternVLA-A1~\citep{cai2026internvla} where it has been evaluated.

\paragraph{Results.} Across all benchmarks, \modelname achieves the best or highly competitive performance. On SimplerEnv (Table~\ref{tab:simplerenv_widowx}), it attains the highest average success rate of 80.8\%, outperforming $\pi_{0.5}$ by 23.7 points. On LIBERO (Table~\ref{tab:libero}), it reaches a 98.9\% average, the best among all methods while ranking first or second on every suite, indicating balanced proficiency across spatial, object, goal, and long-horizon skills. On LIBERO-Plus (Table~\ref{tab:libero_ood}), our zero-shot model obtains the highest total score of 84.8\% and leads on the background, noise, and camera perturbations, demonstrating strong robustness to visual and layout distribution shifts. On RoboTwin~2.0 (Table~\ref{tab:robotwin}), it achieves the best average of 93.2\%, and its accuracy remains nearly unchanged from the clean (93.3\%) to the randomized (93.0\%) setting, confirming robust bimanual coordination under domain randomization. On DOMINO (Table~\ref{tab:domino}), it already surpasses all baselines in the zero-shot setting (27.7\% SR) and improves further to 29.3\% after fine-tuning, reflecting the effectiveness of foresight reasoning for modeling dynamic object interactions. Finally, on EBench (Table~\ref{tab:ebench}), it delivers the best success rate across the Val-Train, Val-Unseen, and Test splits, underscoring its capability on long-horizon indoor mobile manipulation.

\begin{table*}[t]
\centering
\caption{Ablation studies on the foresight reasoning. All experiments are conducted on the two-stage pretrained \modelname.}
\label{tab:ablation}
\resizebox{0.7\linewidth}{!}{%
\begin{tabular}{lcccc}
\toprule
Method & LIBERO & LIBERO-Plus & RoboTwin & DOMINO \\
\midrule
\textbf{\modelname (Ours)} & 98.9 & 84.8 & 93.2 & 27.7 \\
\midrule
\textbf{w/o video loss} & 97.9 & 78.0 & 91.1 & 25.3 \\
\textbf{w/o foresight tokens} & 98.6 & 77.9 & 90.2 & 23.8\\
\bottomrule
\end{tabular}%
}
\end{table*}

\subsection{Analysis}

\paragraph{Ablation Studies.}
We conduct ablation studies to analyze the contribution of the proposed foresight reasoning mechanism.
Experiments are evaluated on four benchmarks, including LIBERO, LIBERO-Plus (zero-shot), RoboTwin, and DOMINO (zero-shot), under the same pretrained model while varying the inference-time configuration.

As shown in Table~\ref{tab:ablation}, removing the video supervision loss leads to consistent performance degradation across all benchmarks.
The drop is more pronounced in zero-shot settings, particularly on LIBERO-Plus and DOMINO.
On LIBERO-Plus, which introduces variations in viewpoint, lighting, and object layout, the gain can be attributed to improved invariance to visual perturbations learned through video-level prediction.
On DOMINO, which requires accurate modeling of highly dynamic object interactions, video supervision further benefits performance by strengthening future-state awareness and motion consistency.
Furthermore, removing the learnable foresight tokens also results in a performance drop across all benchmarks.
This indicates that the foresight tokens serve as a key interface for distilling structured dynamics knowledge from the pretrained video model into the unified expert, enabling effective transfer of spatiotemporal priors for control.

\begin{figure*}[h]
  \centering
  \includegraphics[width=0.35\linewidth]{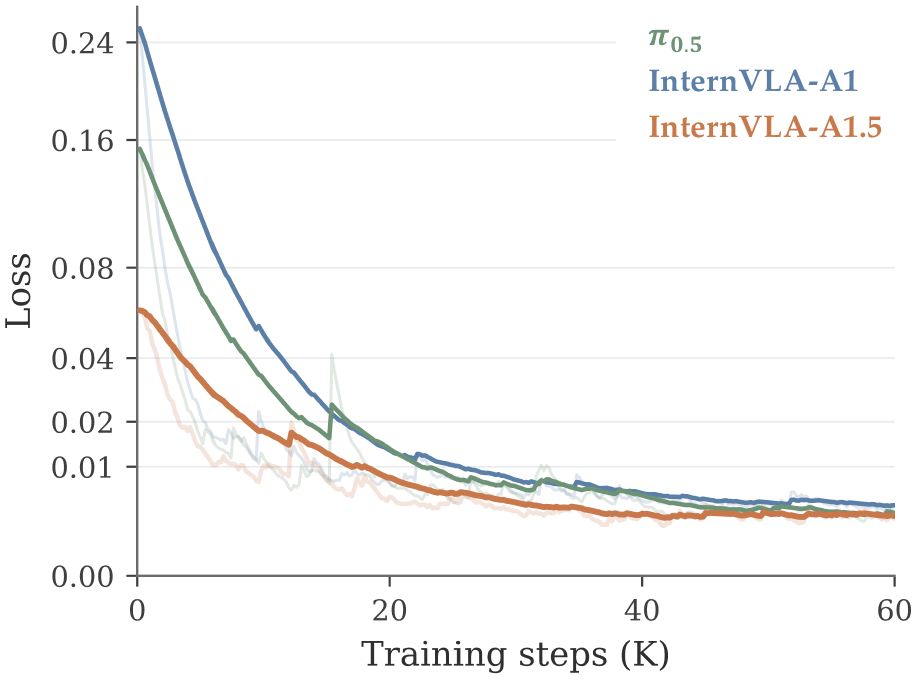}
  \caption{
\textbf{Training efficiency analysis.}
Comparison of SFT loss curves under identical settings.}
\label{fig:loss_curve}
\end{figure*}

\paragraph{Training Efficiency.}
To investigate whether stronger pretraining improves optimization efficiency during downstream adaptation, we compare the SFT loss curves of $\pi_{0.5}$, InternVLA-A1, and \modelname on RoboTwin under the same fine-tuning setup.
All models are trained for 60K steps, corresponding to roughly 1.2 epochs. As shown in Figure~\ref{fig:loss_curve}, \modelname consistently exhibits the fastest convergence and achieves the lowest final training loss.
In particular, while $\pi_{0.5}$ and InternVLA-A1 both start from relatively higher loss regimes and exhibit slower early-stage decay, \modelname reaches a low-loss regime significantly earlier and maintains a more stable optimization trajectory throughout training. This behavior suggests that the representations learned during pretraining induce a more favorable optimization landscape for downstream SFT, effectively reducing the difficulty of policy adaptation.

\begin{figure*}[t]
  \centering
  \includegraphics[width=\linewidth]{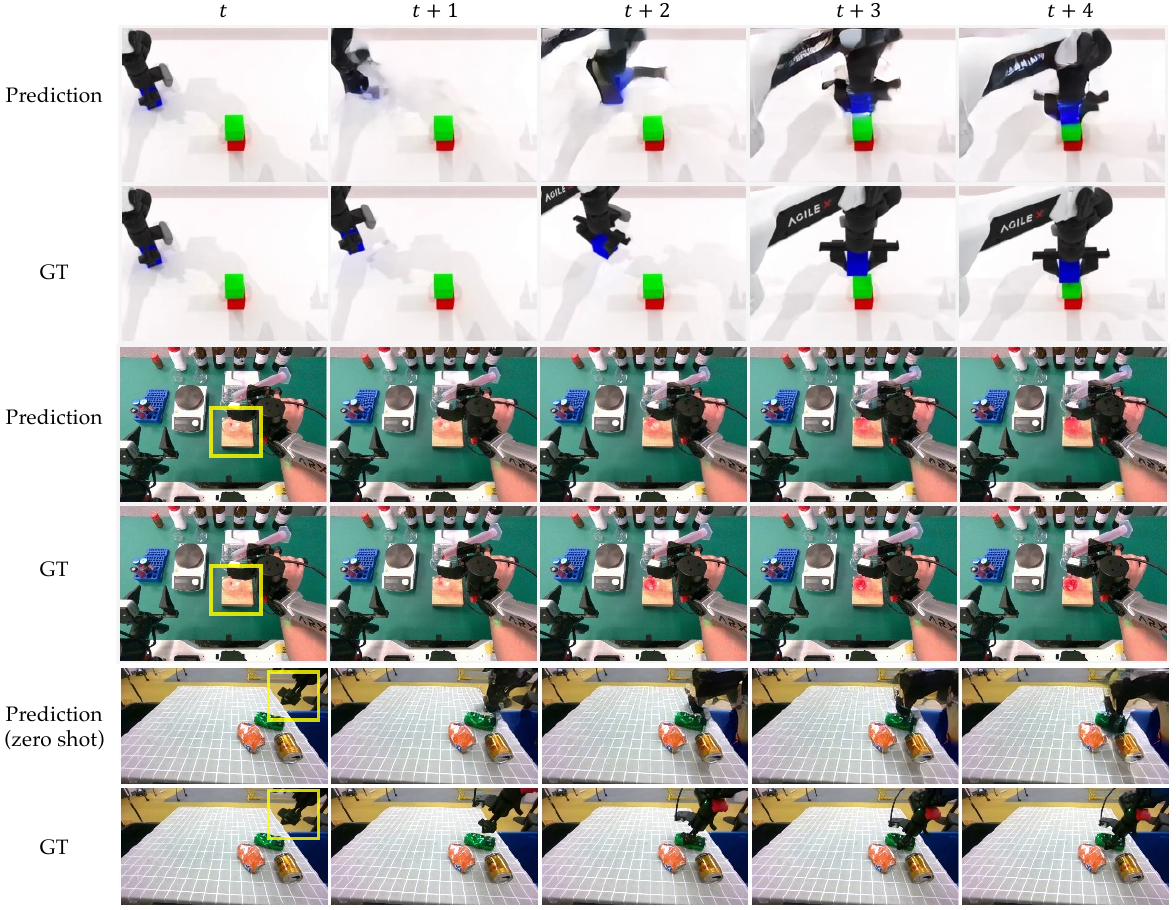}
  \caption{\textbf{Foresight-conditioned future rollouts from the frozen world model.}
  The model predicts future frames ($t \sim t+4$) conditioned on foresight embeddings extracted from the VLM context, demonstrating both accurate motion prediction and physically consistent scene evolution.}
  \label{fig:visualization}
\end{figure*}

\paragraph{World Model Exploitation.}

To study how foresight reasoning leverages pretrained world model priors, we visualize future rollouts generated by the frozen WAN model conditioned on the learned foresight embeddings.
As shown in Figure~\ref{fig:visualization}, the model produces plausible multi-step predictions over $t$ to $t+4$ across both standard evaluation and zero-shot settings.
In the first example, the predictions closely track fine-grained articulated robot motion, indicating that the foresight embeddings effectively capture control-relevant dynamics such as object manipulation trajectories and contact interactions.
In the second example, the model further exhibits consistent physical reasoning by accurately modeling downstream environmental effects, such as changes in liquid level within a container, reflecting an understanding of causal scene evolution beyond pure kinematics.

Notably, even in zero-shot scenarios, the model maintains temporally coherent and physically plausible rollouts, suggesting that the foresight tokens successfully distill spatiotemporal priors from large-scale video pretraining and ground them into action-conditioned world modeling.

\section{Conclusion and Limitations}

In this work, we presented InternVLA-A1.5, a unified framework that integrates vision-language understanding, visual foresight, and action generation on top of a native VLM backbone with a lightweight unified expert. Rather than learning to synthesize future frames, the policy reads out the task-relevant future from its multimodal context through a small set of latent foresight tokens, supervised during training by a frozen pretrained video generation model that is discarded entirely at deployment. Extensive experiments across six simulation benchmarks and four real-world tasks show that InternVLA-A1.5 achieves the best or highly competitive performance, with the clearest gains on zero-shot generalization and long-horizon execution, while keeping a real-time inference speed.

Beyond the results, our study suggests two lessons for building unified models. The first is that prompt design matters. Casting states, control modes, and actions into the native chat template of the VLM, where all targets share one vocabulary and one next-token loss, preserves the pretrained representation, and in our experience this simple choice makes training noticeably more stable and transfers the semantic and instruction-following ability of the VLM into the policy much more fully. The second is that a unified model does not have to learn pixel-level generation from scratch to benefit from future prediction. A handful of latent tokens is enough to encode the future information that action learning needs, so the policy only learns what to imagine while the pretrained generator already carries the knowledge of how the world evolves, and querying this knowledge through a compact latent code comes at a small fraction of the cost of modeling future videos.

\paragraph{Limitations.} Two limitations remain. First, the foresight supervision spans only the short horizon of one action chunk, so the policy absorbs local dynamics priors but does not yet perform long-horizon imagination or explicit planning with the world model. Second, the video generator stays frozen and generic, so the inherited priors are bounded by how well its pretraining covers embodied scenes. We will address these limitations in future work.

\newpage

\bibliography{refs}

\appendix
\newpage

\section{Contributors}
\label{sec:contributors}

Haoxiang Ma$^{*}$, Junhao Cai$^{*}$, Xiaoxu Xu$^{*}$, Hao Li$^{*}$, Yuyin Yang$^{*}$, Yang Tian$^{*}$, Jiafei Cao$^{*}$, 
\mbox{Hongrui Zhu}, Zherui Qiu, Zhaxizhuoma, 
Yuqiang Yang, Jiaqi Peng, Xueyuan Wei, Yangkun Zhu, Jiahao Jiang, 
Xing Gao, Hanqing Wang, Feng Yuan, Kailin Li, Xueyue Zhu, 
Tai Wang, Yan Ding, Jiangmiao Pang, \mbox{Jia Zeng}, 
Jingjing Zhang, Bowen Zhou, Yao Mu, Chunhua Shen, Weinan Zhang$^{\dag}$

{\let\thefootnote\relax
\footnotetext{$^{*}$\,Core contributors. \quad $^{\dag}$\,Corresponding author.}}

\subsection{Real-world Task Details}
\label{app:realworld-tasks}

\begin{figure}[h]
  \centering
  \vspace{-15pt}
  \includegraphics[width=\linewidth]{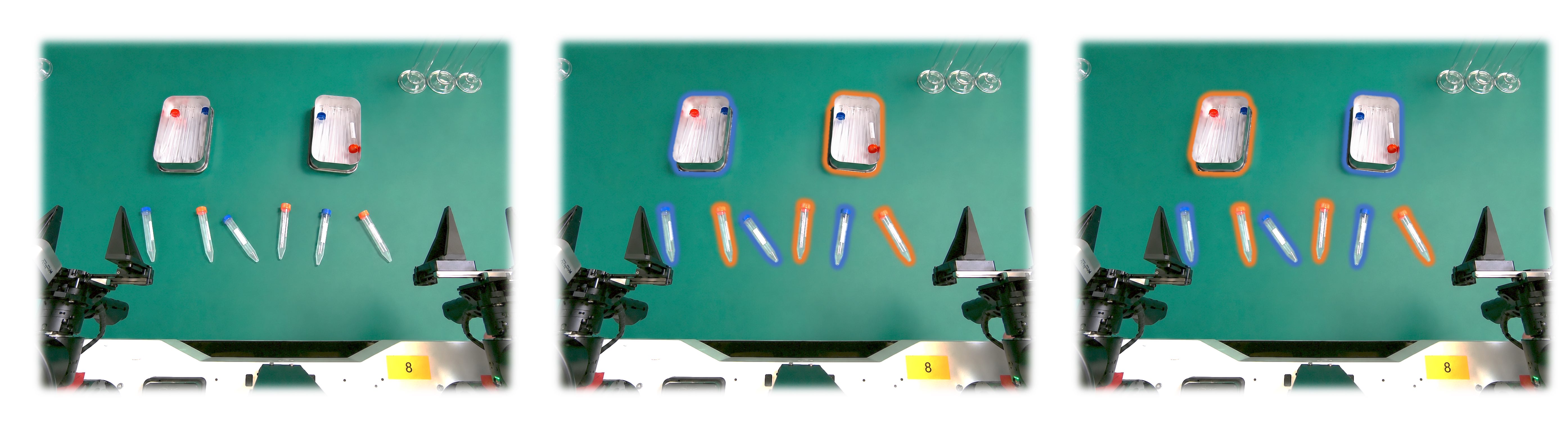}
  \vspace{-20pt}
  \caption{\textbf{Test-tube sorting task. }
    From left to right: (left) the raw scene, containing orange- and blue-tipped plastic tubes and a left and right box; (middle) the seen bindings present in the training demonstrations; and (right) the held-out bindings, evaluated only at test time, where the arm-to-color assignment is reversed. In (middle) and (right), each tube and its target box are drawn in matching outline colors.}
    \label{fig:tube-setup}
    \vspace{-20pt}
\end{figure}

\paragraph{Sort Tubes.} The workspace contains orange- and blue-tipped plastic tubes together with a left and a right collection box, and the robot operates with two arms. Each instruction names an arm and a tube color, such as ``Using the right arm, pick up the orange-tipped plastic tube and place it into the right box'', and the robot uses the named arm to pick a tube of the named color and place it in the box on the same side, so the target box is determined by the chosen arm. There are four (arm, color) bindings in total, and we evaluate two settings. In the full-coverage setting, all four bindings appear in the training demonstrations, which measures how well the policy follows instructions it has already seen. In the held-out setting, the policy is trained only on blue with the left arm and orange with the right arm, and is evaluated on the reversed bindings, blue with the right arm and orange with the left arm, which never appear during training. We run 15 trials per binding under each setting. Figure~\ref{fig:tube-setup} shows the seen and held-out bindings. We choose the held-out bindings so that two shortcuts a policy might otherwise learn both fail on them. A policy that ties each arm to a fixed motion would always grasp blue with the left arm and orange with the right arm, and a policy that ties each color to a fixed arm would always send blue to the left and orange to the right. Both behaviors fit the trained bindings but fail on the held-out ones, so a policy can succeed only by grounding the arm and the color from the instruction independently.

\paragraph{Insert Tubes.} The workspace contains orange- and blue-tipped plastic tubes together with a rack of four holes indexed from 1 to 4, and each instruction names a tube color and a target hole, such as ``Insert the blue-tipped plastic tube into hole 2'', so the robot picks a tube of the named color and inserts it into the specified hole. During training we collect demonstrations for the blue tube at holes 1, 3, and 4 and for the orange tube at holes 1, 2, and 3, so each color covers only a subset of the four holes. At test time we evaluate two settings, an in-domain setting on the (color, hole) bindings seen during training and an out-of-distribution setting on the held-out bindings, namely the blue tube at hole 2 and the orange tube at hole 4, which never appear in the demonstrations. We run 15 trials per binding under each setting, so that success on the held-out bindings requires grounding the color and the target hole from the instruction rather than reproducing a memorized insertion.

\paragraph{Move Tubes.} The workspace contains a left rack and a right rack, each with four holes indexed from 1 to 4, and each instruction names a tube on the left rack and a target hole on the right rack, so the robot picks the specified tube from the left rack and moves it into the specified hole of the right rack. During training we collect demonstrations that move the specific orange tube on the left to holes 1 and 3 on the right and the blue tube on the left to holes 2 and 4 on the right, so each color covers only a subset of the right-side holes. At test time we evaluate two settings, an in-domain setting on the (color, target hole) bindings seen during training and an out-of-distribution setting on the held-out bindings, namely the orange tube to holes 2 and 4 and the blue tube to holes 1 and 3, which never appear in the demonstrations. We run 16 trials per binding under each setting, so that success on the held-out bindings requires grounding the source tube and the target hole from the instruction rather than reproducing a memorized trajectory.

\begin{figure}[h]
    \centering\includegraphics[width=\linewidth]{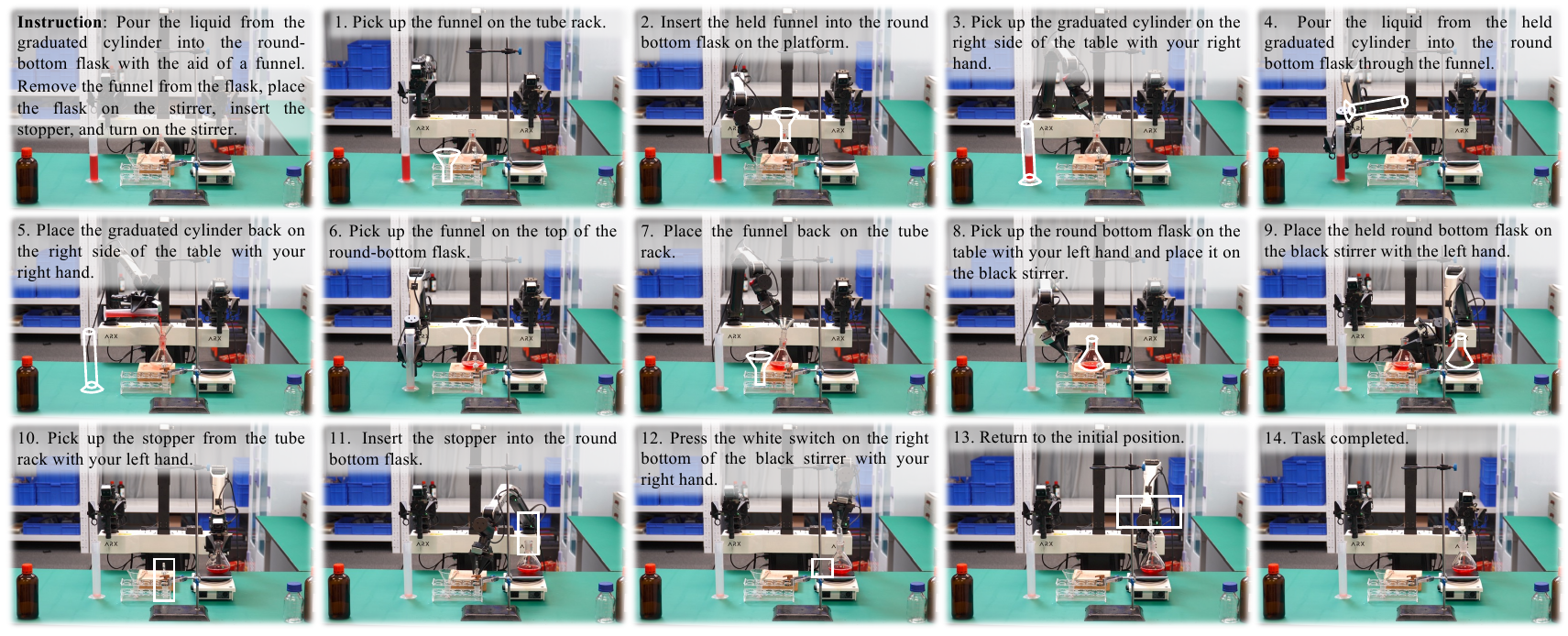}
    \caption{\textbf{The long-horizon MOF task.} The first panel gives the full instruction, and the remaining panels show the 13 sequential subtasks together with the final completed state, from inserting the funnel and pouring the liquid to stoppering the flask and switching on the stirrer. White outlines mark the object or target involved in each step.}
    \label{fig:mof_task}
    \vspace{-20pt}
\end{figure}

\paragraph{MOF.} The task reproduces the solution preparation stage of a metal-organic framework synthesis on a wet-lab bench, and unlike the three instruction-following tasks it contains no held-out bindings and instead targets long-horizon execution. The workspace contains a graduated cylinder holding the liquid on the right side of the table, a round-bottom flask placed on a platform, a funnel and a stopper resting on a tube rack, and a magnetic stirrer, and the robot operates with two arms. A single instruction describes the whole procedure, asking the robot to pour the liquid from the graduated cylinder into the flask with the aid of the funnel, then remove the funnel, transfer the flask onto the stirrer, insert the stopper, and turn on the stirrer. As shown in Figure~\ref{fig:mof_task}, executing the instruction takes 13 sequential subtasks. The robot first fetches the funnel from the rack and inserts it into the flask neck, then grasps the graduated cylinder with the right arm, pours the liquid through the funnel, and places the cylinder back. It then returns the funnel to the rack, carries the flask onto the stirrer with the left arm, retrieves the stopper and seals the flask, presses the switch of the stirrer with the right arm, and finally returns to the initial position. The subtasks must be completed in order, several of them require precise insertion into the narrow flask neck, and pouring changes the scene state, so the policy has to track its progress across stages whose observations differ only in subtle cues such as the liquid level in the flask. We run 20 trials with randomized initial object placements and report the success rate.

\subsection{Simulation Benchmark Details}
\label{app:simbench}


\noindent \textbf{LIBERO.} The LIBERO benchmark~\citep{liu2023libero} contains four task suites, namely Spatial, Object, Goal, and Long, each consisting of 10 tasks with 50 human-teleoperated demonstrations per task, for a total of 2,000 demonstrations. We fine-tune a single model jointly on the mixture of all four suites and evaluate it on each suite separately, reporting the per suite success rate together with their average over 500 rollouts per suite (50 rollouts per task). 

\noindent \textbf{LIBERO-Plus.} LIBERO-Plus~\citep{fei2025libero} augments the LIBERO tasks with systematic visual and layout perturbations to probe robustness under distribution shift. We do not train on LIBERO-Plus; instead, we evaluate the LIBERO checkpoint described above in a zero-shot manner and report the success rate.

\noindent \textbf{RoboTwin 2.0.} RoboTwin 2.0~\citep{chen2025robotwin} consists of 50 bimanual manipulation tasks, each provided under an Easy (clean) and a Hard (domain-randomized) setting. We fine-tune InternVLA-A1.5 on the ALOHA-AgileX embodiment using the full RoboTwin 2.0 training split, which contains 50 clean demonstrations and 500 randomized demonstrations for each task, resulting in 27,500 demonstrations in total. The policy predicts absolute joint actions with an executed action chunk size of 18. We train for 100k steps on 24 GPUs, using a batch size of 16 per GPU. For RoboTwin 2.0, the peak learning rate is set to $1 \times 10^{-4}$ and decays for 140k steps. During evaluation, we report results on both clean and random settings. Specifically, we evaluate 50 tasks in each setting with 100 rollouts per task, resulting in 10,000 evaluation rollouts in total.

\noindent \textbf{DOMINO.} DOMINO~\citep{fang2026towards} is a recently introduced benchmark designed to evaluate whether action models can generalize instructions into precise action execution under dynamic and constrained manipulation scenarios. It introduces a hierarchical motion design together with a continuous Manipulation Score (MS), which measures execution quality beyond binary task completion (Success Rate, SR). Following Qwen-VLA~\citep{wang2026qwen} and PUMA~\citep{fang2026towards}, we evaluate InternVLA-A1.5 on the ALOHA-AgileX embodiment using the 35 Level-1 DOMINO clean setting suites, covering all 35 clean Level-1 tasks with 100 rollouts per task and resulting in 3,500 evaluation rollouts in total. Specifically, we consider two evaluation settings. First, in the zero-shot setting, we directly evaluate the model fine-tuned on RoboTwin 2.0 without any DOMINO training data, which stands for a static-to-dynamic zero-shot evaluation. Second, in the fine-tuned setting, we further fine-tune InternVLA-A1.5 on the DOMINO ALOHA-AgileX Level-1 training split, and then complete a dynamic-to-dynamic evaluation.

\noindent \textbf{EBench.} EBench~\citep{ebench2026} is a benchmark for generalist mobile manipulation, comprising 26 diverse and challenging tasks annotated along 5 capability dimensions and 4 generalization dimensions. Following the standard training protocol in EBench, we fine-tune the pretrained \modelname using a batch size of 128, the AdamW optimizer, and a cosine learning-rate schedule with linear warm-up, with a peak learning rate of $5\times10^{-5}$. Unlike baselines such as $\pi_{0.5}$ and InternVLA-A1, we train for only 100K gradient steps, benefiting from the faster convergence of \modelname.

\noindent \textbf{SimplerEnv.} The SimplerEnv benchmark~\citep{li2025evaluating} evaluates cross-embodiment generalization and the correlation between simulation and real-world performance. Evaluation is conducted on four manipulation tasks. For fine-tuning, we optimize the pretrained \modelname using the AdamW optimizer with a batch size of 128 and a cosine learning-rate schedule with linear warm-up, where the peak learning rate is set to $5\times10^{-5}$.

\end{document}